\documentclass[letterpaper,10 pt, journal, twoside]{IEEEtran}
\usepackage{times}
\usepackage{amsmath,amsfonts}
\usepackage{dsfont}
\usepackage{graphicx}
\usepackage[dvipsnames, svgnames, table]{xcolor}
\usepackage{pifont}
\usepackage{romannum}
\usepackage{makecell}
\usepackage{setspace}
\usepackage{authblk}

\usepackage{multicol}
\usepackage{multirow}
\usepackage[caption=false,font=footnotesize,labelfont=rm,textfont=rm]{subfig}
\usepackage[bookmarks=true]{hyperref}
\usepackage{cleveref}
\usepackage{footnote}
\usepackage{tabularx}
\usepackage{threeparttable}

\crefname{figure}{Fig.}{Fig.}
\crefname{table}{Table}{Table}
\crefname{section}{Sec.}{Sec.}
\crefname{equation}{Eq.}{Eq.}

\newcommand{\yes}{\color{YellowGreen}\ding{51}}
\newcommand{\no}{\color{BrickRed}\ding{55}}

\newcommand{\hl}[1]{#1}

\begin{document}

\title{ScaleADFG: Affordance-based Dexterous Functional Grasping via Scalable Dataset}

\author{Sizhe Wang$^{1,2,3}$, Yifan Yang$^{1,2,3}$, Yongkang Luo$^{1,5,\dag}$, Daheng Li$^{1,2}$, Wei Wei$^{3}$, \\
Yan Zhang$^{1,2}$, Peiying Hu$^{1,2,3}$, Yunjin Fu$^{1,2,3}$, Haonan Duan$^{4}$, Jia Sun$^{1,3}$, Peng Wang$^{1,2,3,\dag}$
\thanks{This paper was recommended for publication by Editor Borràs Sol, Júlia upon evaluation of the Associate Editor and Reviewers' comments.
This work was supported in part by the National Natural Science Foundation of China (Grant 82427802); the China Scholarship Council (Grant 202304910272); and the Beijing Natural Science Foundation Haidian Original Innovation Joint Fund Project (Grants L232035 and L222154).
} 
\thanks{$^{1}$The authors are with the State Key Laboratory of Multimodal Artificial Intelligence Systems, Institute of Automation, Chinese Academy of Sciences, Beijing 100190. China}
\thanks{$^{2}$The authors are with the School of Artificial Intelligence, University of Chinese Academy of Sciences, Beijing 100049, China.}
\thanks{$^{3}$The authors are with the CasiaHand Robotics Co., Ltd, Nanjing 211100, China.}
\thanks{$^{4}$The author is with SenseTime Research, SenseTime, Shanghai 200233, China.}
\thanks{$^{5}$The author is with the Idiap Research Institute, Martigny, Switzerland.}
\thanks{$^{\dag}$Corresponding author: \{yongkang.luo, peng\_wang\}@ia.ac.cn}
}


\maketitle

\begin{abstract}

Dexterous functional tool-use grasping is essential for effective robotic manipulation of tools. However, existing approaches face significant challenges in efficiently constructing large-scale datasets and ensuring generalizability to everyday object scales. These issues primarily arise from size mismatches between robotic and human hands, and the diversity in real-world object scales. To address these limitations, we propose the \textit{ScaleADFG} framework, which consists of a fully automated dataset construction pipeline and a lightweight grasp generation network. Our dataset introduce an affordance-based algorithm to synthesize diverse tool-use grasp configurations without expert demonstrations, allowing flexible object-hand size ratios and enabling large robotic hands (compared to human hands) to grasp everyday objects effectively. Additionally, we leverage pre-trained models to generate extensive 3D assets and facilitate efficient retrieval of object affordances. Our dataset comprising five object categories, each containing over 1,000 unique shapes with 15 scale variations. After filtering, the dataset includes over 60,000 grasps for each 2 dexterous robotic hands. On top of this dataset, we train a lightweight, single-stage grasp generation network with a notably simple loss design, eliminating the need for post-refinement. This demonstrates the critical importance of large-scale datasets and multi-scale object variant for effective training. Extensive experiments in simulation and on real robot confirm that the ScaleADFG framework exhibits strong adaptability to objects of varying scales, enhancing functional grasp stability, diversity, and generalizability. Moreover, our network exhibits effective zero-shot transfer to real-world objects. Project page is available at \href{https://sizhe-wang.github.io/ScaleADFG_webpage}{\url{https://sizhe-wang.github.io/ScaleADFG_webpage}}

\end{abstract}

\begin{IEEEkeywords}
Functional Grasping; Dexterous Manipulation; Deep Learning in Grasping and Manipulation; Affordance
\end{IEEEkeywords}

\IEEEpeerreviewmaketitle

\section{Introduction}
\vspace{-1mm}

\begin{figure}[tbp]
    \centering
    \includegraphics[width=1.0\linewidth]{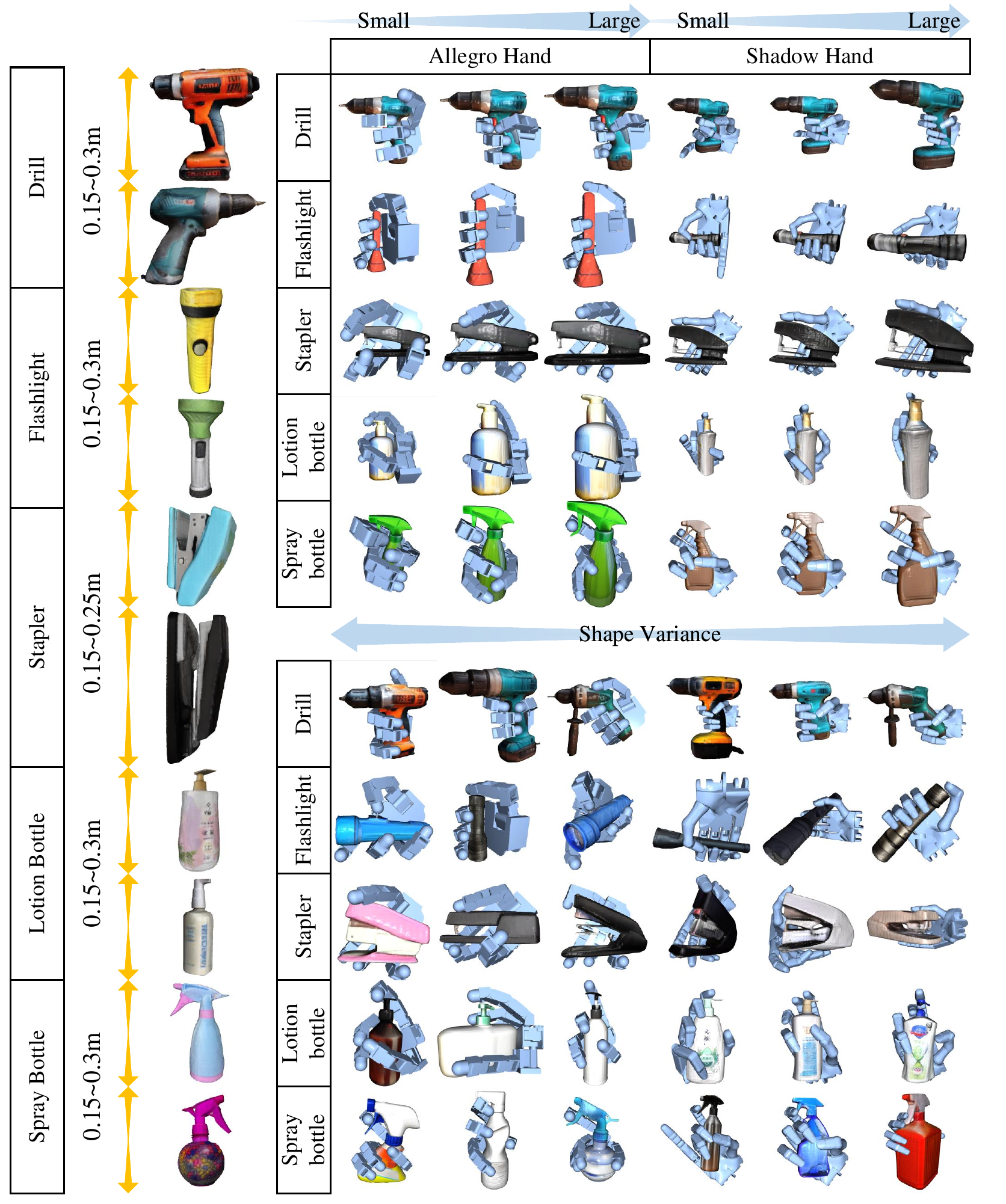}
    \caption{Adaptability on scale variance and shape variance for both hands.
}
    \label{fig:higlight}
    \vspace{-2mm}
    \vspace{-4mm}
\end{figure}

\begin{figure*}[!tbp]
    \centering
    \includegraphics[width=\linewidth]{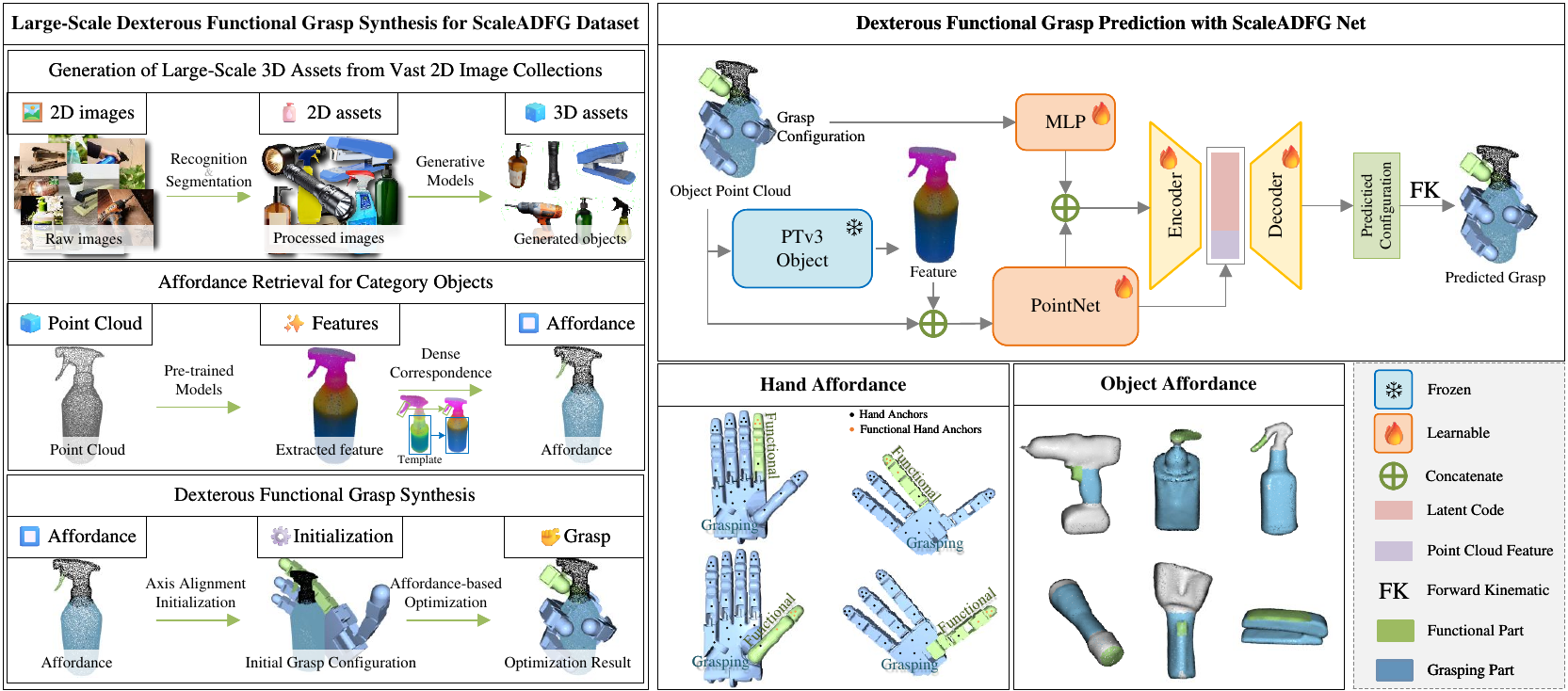}
    \caption{
    Overview of the ScaleADFG pipeline for dexterous functional grasping. The large-scale \textbf{ScaleADFG-Dataset} comprises automated 3D object generation from internet images, affordance retrieval using dense correspondence, and optimization-based synthesis of functional grasps with axis alignment initialization. A light weight conditional variational autoencoder (CVAE) based \textbf{ScaleADFG-Net} predicts grasps for diverse object shapes and scales, trained with the constructed dataset. \hl{In real world, we use the ScaleADFG-Net to inference grasp configurations.} Hand and object affordances are illustrated, with matching colors indicating the alignment of functional parts and grasping parts.
    }
    \label{fig:framework}
    \vspace{-2mm}
    \vspace{-4mm}
\end{figure*}

\IEEEPARstart{T}{he} ability to perform dexterous functional grasping is a cornerstone of robotic manipulation, underpinning a wide spectrum of applications from domestic service to precision industrial tasks. Unlike universal grasping, which primarily focuses on stability, functional grasping requires the synthesis of grasp configurations that enable the execution of specific post-grasp manipulations, integrating both object affordances and task intent. Achieving such capabilities in robotic hands, particularly those with high degrees of freedom, remains a formidable challenge due to the scarcity of large-scale, diverse, and semantically rich dataset.

Recent advances in deep learning have significantly improved the generalizability of robotic grasping strategies. However, the construction of large-scale datasets for dexterous functional grasping is still hindered by several factors. Previous methods often rely on human hand demonstrations, which are labor-intensive and limited in quantity. These demonstrations are typically retargeted to robotic hands either directly \cite{mandikal2022dexvip, qin2022dexmv} or via contact map alignment \cite{ContactGrasp, dexfg}, but such approaches inherit the biases of human hand size and kinematic, resulting in datasets skewed toward larger objects and less precise functional grasps. Moreover, the dependence on existing object datasets or manual 3D scanning further restricts scalability and diversity.

A promising alternative lies in affordance-based grasp synthesis. The concept of affordance, first introduced by Gibson \cite{Gibson_affordance_1979}, refers to the actionable properties of object parts that suggest potential interactions. By reasoning about object affordances rather than imitating human demonstrations, robots can develop more intuitive, adaptive, and generalizable grasping strategies. Affordance-based approaches also align more closely with human cognitive processes in object interaction, providing a more fundamental logic for grasp planning.

Furthermore, affordance-based algorithms are well-suited to exploit the capabilities of pretrained foundation models. By utilizing segmented images from vision models and abundant internet image resources, we can efficiently generate large numbers of 3D object assets through generative models. Affordance parts of these 3D objects can then be identified using features extracted from pretrained perceptual models, supporting scalable and automated dataset construction and labeling.

In this work, we propose \underline{Scalable} \underline{A}ffordance-based \underline{D}exterous \underline{F}unctional \underline{G}rasping generation framework \textbf{ScaleADFG}. The framework comprises two main components: 
(1) A large-scale functional grasp dataset \textbf{ScaleADFG-Dataset} constructed via an affordance-based synthesis algorithm with pretrained models, and 
(2) A single-stage grasp generation network (\textbf{ScaleADFG-Net}) trained jointly across object categories. The network exhibits zero-shot transfer from synthesized objects to real world objects. 
Our contributions can be summarized as follows:
\begin{enumerate} 
    \item We propose the ScaleADFG pipeline, an efficient and scalable grasp generation framework leveraging pretrained foundation models for image segmentation, 3D reconstruction, and functional affordance perception. This framework automates dataset generation, significantly reducing reliance on expert demonstrations while maintaining high semantic consistency and functional relevance.
    \item We introduce a comprehensive dexterous functional grasp dataset, characterized by extensive diversity in object scales and shapes. Generated through an affordance-based synthesis algorithm with a modular and replaceable hand kinematic component, this dataset encompasses over 1,000 object shapes per category, each presented in 15 distinct scale variations for 2 robotic hands (Shadow Hand and Allegro Hand), thus facilitating robust dexterous grasp learning.
    \item We develop a lightweight, CVAE-based grasp generation network with a notably simple loss design, capable of generalizing across all object categories within a unified model. Trained exclusively on our synthetic dataset, the network exhibits strong zero-shot transfer capabilities to real-world objects, validated through extensive experiments in simulation and real robot settings.
\end{enumerate}

\section{Related Work}

\subsection{Dexterous Grasp Dataset}
Current grasp synthesis methodologies are broadly categorized into analytical \cite{millerGraspitVersatileSimulator2004} and data-driven paradigms \cite{wangDexGraspNetLargeScaleRobotic2023,  lundellMultiFinGANGenerativeCoarseToFine2021, duan2022learning, duanReactiveHumantoRobotDexterous2025}. The analytical approach includes sampling and optimization methods. Graspit! \cite{millerGraspitVersatileSimulator2004} synthesis dexterous grasps from sampling linear combination of eigen grasps. ContactGrasp \cite{ContactGrasp} filter dexterous functional grasp from Graspit! \cite{millerGraspitVersatileSimulator2004} according to contact map. By sampling from eigen grasp, the dimensionality of the sampling space can be reduced, but the diversity of grasp is also reduced. Optimization-based methods have become more prevalent in recent research. Dexgraspnet \cite{wangDexGraspNetLargeScaleRobotic2023} synthesis a large-scale dexterous universal grasp dataset based on a differentiable froce-clousure metric in \cite{force_closure2022zhusongchun}. Functional grasp needs additional semantic labels, whose collection is time-consuming. OakInk \cite{yang2022oakink} transfers hand-object contact map from a template to whole category based on DeepSDF\cite{deepsdf_2019_CVPR} and construct a large-scale functional grasp dataset with semantic label for human hand. \cite{dexfg} constructs a functional grasp dataset for multiple dexterous hands using similar transfer method.

Large-sacle dexterous functional dataset is limited. This article undertakes a comprehensive investigation of large-scale dataset construction through automated object generation and affordance and grasp annotation framework. By harnessing pre-trained foundation models, our proposed system enables systematic 3D object generation from the Internet images, paired with affordance part and functional grasp annotation. 

\subsection{Dexterous Functional Grasping} \label{Structural_Representations}
Dexterous functional grasping \cite{fang2020learning, zhu2021toward, yang2022oakink, dexfg,mandikal2022dexvip, dapg, yang2021cpf} focuses on synthesizing grasp configurations that not only ensure stability but also enable the execution of task-specific manipulations. \cite{dapg} use Deep Reinforcement Learning (DRL) to learn dexterous functional grasping. \cite{mandikal2022dexvip} generate functional grasp for dexterous hands by imitate human hand grasp directly. Some works \cite{ContactGrasp, dexfg} assume that the functional grasp of human hand and dexterous hands share a common contact map. Gibson \cite{Gibson_affordance_1979} proposed the concept of affordance, which describes the object property for interacting with the external environment. Functional grasping generated based on affordance may be more aligned with human cognition in grasping and is a more basic underlying logic in grasping. However, there are limited work generate dexterous functional grasp with object affordance part. Kokic et al. \cite{kokic2017affordance_grasp} utilize affordance part generates functional grasp for a three finger gripper.

Generate dexterous functional grasp based on object affordance part has potential to get rid of human demonstrations and turn to rely on associations between hand affordance parts and object affordance parts. The association can be defined by rules. In this work, we explore this direction by employing such a method. As a result, we construct a large-scale dexterous functional grasp dataset without human grasping demonstrations. Additionally, we found that this method also relaxes the constraints on the relative scale between the hand and the object.

\section{Large-Scale Dexterous Functional Grasp Synthesis for ScaleADFG Dataset} \label{sec:dataset}
\subsection{Generation of Large-Scale 3D Assets from Vast 2D Image Collections}
A large set of images was collected through Internet searches using relevant keywords. These images were filtered and segmented using Grounded SAM 2 \cite{renGroundedSAMAssembling2024, ravi2024sam2segmentimages}. 3D object assets were generated from the 2D images using multiple pre-trained generative models, including InstantMesh \cite{xu2024instantmesh}, TripoSR \cite{tochilkin2024triposr}, and TRELLIS \cite{xiang2024trellis}. The three generative models were randomly selected to promote mesh diversity.

To filter out flawed outputs from the pre-trained models, the following steps were carried out: (1) Irrelevant images resulting from segmentation were manually removed through a rapid thumbnail screening process, which required approximately 10 minutes per category; (2) Meshes falling outside the file size range of 1M–10M were discarded to eliminate degenerate geometries, such as unstructured surfaces or poor-quality mesh structures; (3) The remaining meshes were further filtered based on grasp quality metrics, \hl{including $d_G \leq 0.02m, d_F \leq 0.002m, d_{IP} \leq 0.002m, d_{SP} \leq 0.002m$, which are defined in} \cref{sec:metrics}. 
Successful and high-quality grasps and their corresponding meshes were retained even when abnormalities (e.g., incomplete meshes, extra components, irrelevant parts, or hallucinated structures) were present, \hl{as preserving such abnormal meshes increases dataset complexity and enhances robustness to complex or imperfect real-world cases.}

\subsection{Affordance Retrieval for Category Objects}
\label{sec:part_annotation}

Pre-trained models integrated into the affordance retrieval process are used to enhance generalization and efficiency. We employed the pre-trained PTv3-object model from SAMPart3D \cite{yangSAMPart3DSegmentAny2024} as the feature extractor for object point clouds. This model distills visual features from FeatUp-DINOv2 \cite{fu2024featup, oquabDINOv2LearningRobust2023}. \hl{These models learn highly generalized and semantically consistent feature representations from vast data, making the features inherently robust to intra-category shape variations.}

As noted in \cite{oquabDINOv2LearningRobust2023}, \hl{DINOv2 features exhibit the property that components corresponding to object ``parts'' match well across images of the same category.} The 3D features extracted by PTv3-object in SAMPart3D \cite{yangSAMPart3DSegmentAny2024}, which are distilled from DINOv2, are expected to inherit this property, effectively lifting it to 3D point clouds. 
\hl{This robustness ensures that dense correspondences are established reliably across different instances of a category, even when their shapes differ significantly. Consequently, the affordance propagation does not rely heavily on the particular choice of the template, which minimizes the risk of introducing bias at the category level.}

Building on this, we identified affordance parts through dense feature matching across objects within the same category. Specifically, a single random object template per category was manually annotated with affordance part labels, and corresponding parts in other instances were determined by matching to this template. This annotation process remained lightweight while \hl{benefiting from the strong correspondence capabilities of the extracted features}, which also remain reliable under significant rotational variations of the point clouds.

\subsection{Dexterous Functional Grasp Optimization} \label{sec:grasp_synthesis}

We employed an affordance-based strategy, as illustrated in \cref{fig:framework}, to generate dexterous functional grasps by identifying two key affordance parts: the functional part and the grasping part on both the object and the hand. During synthesis, corresponding affordance parts were encouraged to remain in close proximity to ensure both functional performance and grasp stability. Furthermore, physical plausibility was enforced by minimizing hand-object and hand self-penetration. With these considerations, the dexterous functional grasps generated by our method were functional, stable, and physically plausible.

A significant contribution of our approach is its ability to handle object scale variations. We specifically examined common real-world sizes for each category and defined a scale range from $S_\text{low}$ to $S_\text{high}$ for all object categories. This scale is determined by the maximum extent of the object's oriented bounding box. Within this range, we sampled 15 scales to synthesize grasps. Objects at the extremes of this spectrum, especially at smaller sizes, posed significant challenges. The relatively larger size of the robotic hand compared to a human hand increased the risk of self-collisions and often led to unrealistic grasps for small objects. To address these challenges, our method adapts by changing gestures rather than slightly adjusting finger configurations, as is typical in contact map-based approaches. This design enables robust generation of dexterous functional grasps even for small-scale objects.

\subsubsection{\textbf{Axis Alignment Initialization}}
\begin{figure}[!tbp]
    \centering
    \includegraphics[width=0.7\linewidth]{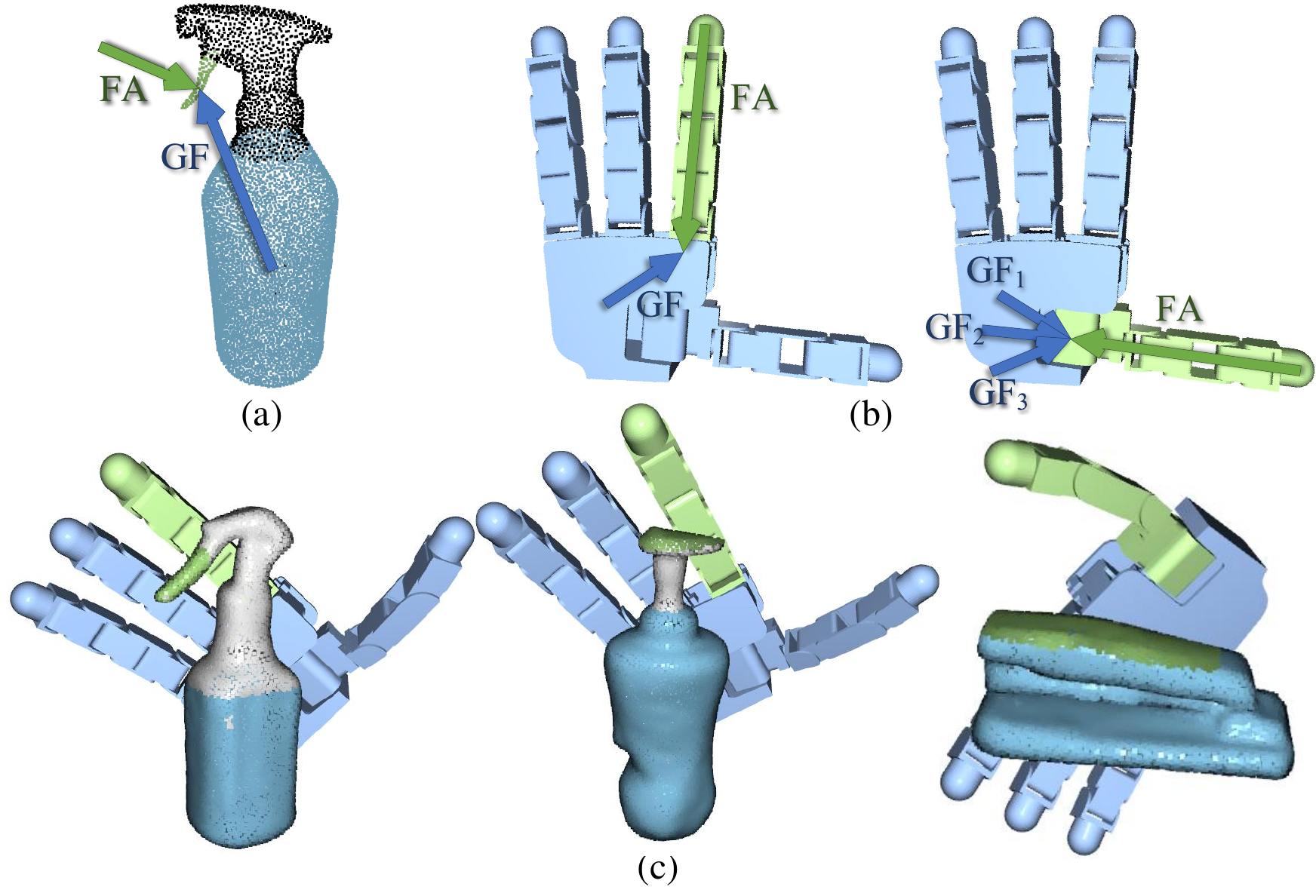}
    \vspace{-3mm}
    \caption{Illustration of the proposed initialization method. The object (a) and hand (b) are each associated with two axes, with corresponding axes indicated by the same color to show the alignment between grasping and functional parts. (c) presents examples of the initialization results.
}
    \label{fig:init}
    \vspace{-2mm}
    \vspace{-4mm}
\end{figure}

The performance of optimization-based grasp synthesis is highly sensitive to the quality of initialization. To address this, we proposed an axis alignment-based initialization strategy to enhance functional dexterous grasping, as illustrated in \cref{fig:init}, using the case where the index finger served as the primary functional finger. In our approach, both the hand and the object are assigned two types of axes: the Grasp-to-Functional Axis ($GF_i$) and the Force-Application Axis ($FA_i$). For the object, these axes are defined based on the relative positions and surface normals of the grasping region and the functional part. For the hand, the axes are derived from vectors connecting the grasp center, functional contact region, and fingertips. Notably, to account for the thumb’s higher degree of mobility, three separate Grasp-to-Functional Axes are defined for it according to \cite{grasp_taxonomy}. The initialization process proceeds in two stages. First, the primary axes ($GF_i$) of the hand and object are aligned precisely. Next, the secondary axes ($FA_i$) are aligned within the plane orthogonal to the primary axis to ensure appropriate directional force application. After orientation alignment, the hand was translated towards the object until initial contact was achieved without penetration, ensuring a collision-free starting configuration. Finally, the finger joint angles are initialized in a slightly flexed inward posture to better support subsequent optimization.

\subsubsection{\textbf{Affordance-based Optimization}}

The goal of dexterous functional grasp synthesis is to synthesize an optimal functional grasp configuration $\boldsymbol G^*$ that satisfies the requirements of functionality, stability, and physical plausibility when given an object point cloud $\mathcal P_O$, by minimizing the following losses. In addition, the functional part of the object for tool use $\mathcal A_{O_F}$ and the stable grasping part $\mathcal A_{O_G}$, annotated in \cref{sec:part_annotation}, served as supervisory signals for the loss functions.

\paragraph{Functional Part Distance Loss} The functional part is defined as the region where the intended manipulation is performed, such as the drill button, and is represented as a point cloud $\mathcal A_{O_F}$. The functional part distance loss selects specific hand anchor points $\mathcal A_{H_F}$ in \cref{fig:framework} according to the designated hand functional part, and computes the distances between $\mathcal A_{O_F}$ and $\mathcal A_{H_F}$. This loss encourages the corresponding affordance parts to move closer. The formulation is given as follows:
\begin{equation}
    L_\text{F} = CD\left(\mathcal A_{H_F}, \mathcal A_{O_F}\right)
\end{equation}
where $CD$ denotes the Chamfer Distance, defined as:
\begin{equation}
    \begin{split}
    CD(\mathcal{P}, \mathcal{Q}) = &\frac{1}{N_{\mathcal{P}}} \sum_{\boldsymbol p \in \mathcal{P}} \min_{\boldsymbol q \in \mathcal{Q}} \|\boldsymbol p - \boldsymbol q\|_2^2 \\
    + &\frac{1}{N_{\mathcal{Q}}} \sum_{\boldsymbol q \in \mathcal{Q}} \min_{\boldsymbol p \in \mathcal{P}} \|\boldsymbol q - \boldsymbol p\|_2^2
    \end{split}
\end{equation}
where $\mathcal{P}, \mathcal{Q}$ denotes two point clouds, and $\boldsymbol p, \boldsymbol q$ are points belongs to $\mathcal{P}, \mathcal{Q}$ respectively. $N_{\mathcal{P}}, N_{\mathcal{Q}}$ are the number of points in $\mathcal{P}, \mathcal{Q}$ respectively.

\paragraph{Grasping Part Distance Loss}

The grasping part is typically located at the handle or body of the object, such as the handle of a drill, and is represented as the point cloud $\mathcal A_{O_G}$. The grasping part distance measured the proximity of the corresponding hand part to the object’s grasping part. The loss function is formulated as follows:

\begin{equation}
    L_\text{G} = CD\left(\mathcal A_{H_G}, \mathcal A_{O_G}\right)
\end{equation}

where $\mathcal A_{H_G}$ represents the anchor points on the palm and fingers within the hand’s grasping part.

\paragraph{Force Closure Loss}

Following \cite{force_closure2022zhusongchun,wangDexGraspNetLargeScaleRobotic2023}, we incorporated a force closure loss $L_\text{FC}$ to promote grasp stability. This loss measures the deviation from ideal force–torque equilibrium and the violation of friction constraints, ensuring that the applied forces and torques prevent slippage or instability. Given $n$ contact points $x_i \in \mathbb{R}^3$ and their corresponding friction cone axes $c_i$, the loss is formulated as:
\begin{align}
    L_{FC} = \left|\left|Gc\right|\right|_2
\end{align}
where
\begin{align}
    G=&
    \begin{bmatrix}
         I_{3\times3} & ... & I_{3\times3}\\
         [x_1]_\times  & ... & [x_n]_\times 
    \end{bmatrix} \\
    [x_i]_\times=&
    \begin{bmatrix}
         0 & -x_i^{(3)} & x_i^{(2)}\\
         x_i^{(3)} & 0 & -x_i^{(1)}\\
         -x_i^{(2)} & x_i^{(1)} & 0
    \end{bmatrix} 
\end{align}

\paragraph{Hand-Object Interpenetration Loss}

In consideration of generating physically plausible grasp, we minimized the hand-object interpenetration. The loss is formulated as following:

\begin{equation}
    L_\text{IP}=\sum_{\mathcal H_{p_i} \in \mathcal H} -\min\left[SDF\left(\mathcal{P_O} | \mathcal H_{p_i}\right), 0\right]
\end{equation}

where $SDF\left(\mathcal P|\mathcal Q\right)$ denotes the Signed Distance Field (SDF), which represents the distance to the nearest surface in a geometric space of $\mathcal Q$, with the sign indicating whether the point set $\mathcal P$ is inside ($-$) or outside ($+$) the surface. $\mathcal P_O$ denotes object point cloud, $\mathcal H$ denotes the whole hand mesh, and $\{\mathcal H_{p_i} \in \mathcal H\}$ denote parts of hand mesh including fingers and palm.

\paragraph{Hand Self Penetration Loss}

Hand self penetration is computed as distances between each pair of fingers. The loss is formulated as following:

\begin{equation}
    L_\text{SP} = \sum_{i}\sum_{j\ne i}- \min\left[\mathop{SDF}\left(\mathcal P_{H_{p_j}} | \mathcal H_{p_i}\right), 0\right]
\end{equation}

where $\mathcal H_{p_i}$ and $\mathcal H_{p_j}$ denotes different parts of hand including fingers and palm, and $\mathcal P_{H_{p_i}}$ denotes surface point cloud of $\mathcal H_{p_i}$.

\paragraph{Total Loss}

For the above loss, we calculated the weighted sum of each loss mentioned above as total loss, as shown in \cref{eq:total_loss}:

\begin{equation} \label{eq:total_loss}
    \begin{split}
        L_\text{total}=&\lambda_\text{F}L_\text{F}+\lambda_\text{G}L_\text{G}+\\
        &\lambda_\text{FC}L_\text{FC}+\lambda_\text{IP}L_\text{IP}+\lambda_\text{SP}L_\text{SP}
    \end{split} 
\end{equation}
where $\lambda_i$ are trade-off weights to balance each loss.

Additionally, to ensure the high quality of the grasp dataset, we filtered the synthesised grasps based on metrics in \cref{sec:metrics} \hl{($d_G \leq 0.02m, d_F \leq 0.002m, d_{IP} \leq 0.002m, d_{SP} \leq 0.002m$)} to retain those of higher quality. Following DexFG \cite{dexfg}, we also implemented a refinement and filtering process with the simulation environment, removing grasps with severe collisions and retaining only the successful grasps.

With our proposed initialization method, grasp optimization starts from a configuration close to the optimal dexterous functional grasp. This significantly reduces the number of iterations required and lowers the risk of convergence to suboptimal local minima, improving both efficiency and grasp quality. Our method converges in fewer than 200 steps. Leveraging batch optimization, we can synthesize 150 dexterous functional grasps in under 30 seconds on a single RTX 3090 Ti GPU, averaging less than 0.2 seconds per grasp. This matches DexFG \cite{dexfg} (0.25s per grasp) and is significantly faster than ContactGrasp \cite{ContactGrasp}, which takes over 0.5 hours per grasp.

\section{Dexterous Functional Grasp Prediction with ScaleADFG Net}
\label{sec:network}
ScaleADFG employs a lightweight network architecture to predict dexterous functional grasps, avoiding the large computational workload of complex models and showing generalizability and high grasp quality at the same time.

The proposed network as illustrated in \cref{fig:framework} adopts a lightweight Conditional Variational Autoencoder (CVAE) \cite{kingmaAutoEncodingVariationalBayes2022} architecture, integrating pretrained object feature extractor PTv3 Object from SAMPart3D \cite{yangSAMPart3DSegmentAny2024} and PointNet \cite{charlesPointNetDeepLearning2017} for efficient grasp prediction. Object point clouds and the first three principal components of the extracted features are concatenated and processed by the encoder, which maps them to a latent space. The decoder predicts grasp configurations, which are then realized via differentiable forward kinematics. Our training employs two loss functions:

    \subsubsection{Reconstruction Loss} Align the point cloud of predicted grasp $\hat{\mathcal{P}}_H$ and the ground truth $\mathcal{P_H}$, where $CD$ denotes Chamfer Distance. 
    \hl{We adopt a point cloud–based reconstruction loss instead of supervising joint angles, as it simultaneously constrains joint angles, rotations, and translations, thereby reducing sensitivity to hyperparameters.}

        \begin{equation} \label{eq:lrec}
            L_\text{Rec}=CD{\left({\hat{\mathcal P}}_H, \mathcal{P_H}\right)}
        \end{equation}
        
    \subsubsection{KL Divergence} Regularizes the latent space to follow a Gaussian distribution $\mathcal N\left(\boldsymbol 0,\boldsymbol I\right)$, promoting generalizability.

        \begin{equation} \label{eq:kld}
            \begin{split}
                L_\text{KLD}&=KL\left(\mathcal N\left(\boldsymbol\mu,\boldsymbol\Sigma\right)||\,\mathcal N\left(\boldsymbol 0,\boldsymbol I\right)\right)\\
                &=\frac{1}{2}\sum_{i=1}^{N_z}\left(\mu_i^2+\sigma_i^2-\ln\sigma_i^2-1\right)
            \end{split}
        \end{equation}
        where $N_z$ denotes the hidden dimension of the autoencoder, $\boldsymbol\mu=\begin{bmatrix}\mu_1&\mu_2&\cdots&\mu_{N_z}\end{bmatrix}$ and $\boldsymbol\Sigma=\begin{bmatrix}\sigma_1&\sigma_2&\cdots&\sigma_{N_z}\end{bmatrix}$ are predicted by the encoder.

At inference, the encoder is removed and latent codes are sampled from $\mathcal{N}(\boldsymbol{0}, \boldsymbol{I})$, allowing the network to generate functional grasps for novel objects and scales. This enables generalization and zero-shot transfer from synthetic to real-world scenarios.

\section{Experiments} \label{sec:experiment}

\begin{table}[!tbp]
    \caption{Dexterous Grasping Dataset Comparison}
    \label{tab:dataset_comparison}
    \renewcommand{\arraystretch}{1.2}
    \centering
    
    \resizebox{1.0\linewidth}{!}{
    \begin{threeparttable}
    
        \begin{tabular}{c|c|c|c|c|c}
            \Xhline{1.2pt}
            Dataset      & $\overline{ N_\text{obj}/N_\text{Cat.}}$   & $N_\text{grasp}$    & Functional & Multi-Hands  & Scale Variety        \\
            \hline

            DexGraspNet \cite{wangDexGraspNetLargeScaleRobotic2023} & $\approx$\,41    & \textbf{1.32M}     & \no    & \no   & \no               \\

            ContactDB \cite{contactdb}  & 1 & 1,350  & \yes & \no & \no\\

            Zhu et al. \cite{zhu2023toward}  & $\approx$\,8 & 13k & \yes  & \no & \no \\

            OakInk \cite{yang2022oakink}  & $\approx$\,57 & 50k  & \yes  & \no & \no\\

            AffordPose \cite{jianAffordPoseLargescaleDataset2023}  & $\approx$\,50 & 26k  & \yes  & \no & \no \\

            DexFuncGrasp \cite{hangDexFuncGraspRoboticDexterous2024}  & $\approx$\,47 & 14k  & \yes  & \no & \no\\

            DexFG \cite{dexfg}  & \underline{$\approx$\,63} & 10k  & \yes & \yes & \no\\
            \hline
            ScaleADFG (Ours)  & \textbf{$\approx$\,1.3k} & \underline{65k}  & \yes  & \yes & {\yes}\\
            \Xhline{1.2pt}
            
        \end{tabular}
    
    \begin{tablenotes}
    \footnotesize
    \item $N_\text{Cat.}$, $N_\text{obj}$, and $N_\text{grasp}$ represent the number of categories, objects, and grasps respectively. 
    \item \textbf{Bold} means the best; \underline{Underline} means second-best.
    \end{tablenotes}
    \vspace{-8mm}
    \end{threeparttable}
    
    }
\end{table}

\subsection{Metrics}
\label{sec:metrics}

\begin{table}[!bp]
\vspace{-2mm}
\vspace{-4mm}
    \caption{Evaluation for Dataset Quality}
    \label{tab:human_study}
    \renewcommand{\arraystretch}{1.2}
    \centering
    \resizebox{\linewidth}{!}{
        \begin{tabular}{c|l|ccccc|c}
            \Xhline{1.2pt}

            \multicolumn{2}{c|}{Category} & Drill & Flashlight  & Stapler & Spray Bottle & Lotion Bottle & Avg. \\
            \hline
            \multirow{5}{*}{\makecell{Human\\Study}}& Avg. Abnormal Shape Ratio $\downarrow$     & 4/50 & 4/50  &  4/50 & 1/50 & 1/50 & 5.60\%  \\
            & Avg. Functional Part Error Ratio $\downarrow$      & 2/50 & 1/50 &  9/50 & 1/50 & 5/50 & 7.20\%  \\
            & Avg. Grasping Part Error Ratio $\downarrow$      & 1/50 & 2/50 & 0/50 & 0/50  & 2/50 & 2.00\%  \\
            & Avg. Stable Ratio $\uparrow$      & 45/50 & 43/50 &  45/50 & 48/50 & 48/50 & 91.60\%  \\
            & Avg. Functional Ratio $\uparrow$  & 45/50 & 40/50 &  35/50 & 46/50 & 37/50 & 81.20\%\\
            \hline
            \multirow{2}{*}{\makecell{Affordance\\Accuracy}}& Functional Part $\uparrow$ & 92.84\% & 92.40\% & 85.76\% & 93.32\% & 91.27\% & 91.12\% \\
            & Grasping Part $\uparrow$ & 91.76\% & 81.15\% & 98.29\% & 86.83\% & 80.42\% & 87.69\% \\
            \Xhline{1.2pt}
        \end{tabular}
    }
\end{table}

\subsubsection{\textbf{Success Rate (SR)} for Stability}\label{sec:sr}

Experiments are conducted in IsaacSim \cite{mittalOrbitUnifiedSimulation2023}. Objects (1 kg) are initialized at the origin under zero-gravity, and the robotic hand is set to the target grasp configuration. To enhance stability, 0.5 radians is added to the flexion of inward-bending joints. $6$ sequential force application phases are performed, each applying a $10\,\text{N}$ force along the $\pm x$, $\pm y$, and $\pm z$ axes. A grasp is considered successful if the object remains held throughout all phases without dropping. Notably, such criteria is significantly more stringent than many of those adopted in prior works, which often assess under gravity alone and with much lighter objects.

\subsubsection{\textbf{Grasping Part Distance ($\boldsymbol{d_\text{G}}$)} for Contact}
Measures the alignment between the hand's grasping part $\mathcal A_{H_G}$ and the object's grasping part $\mathcal A_{O_G}$, where $CD$ is Chamfer Distance:
\begin{equation} \label{eq:dg}
        d_\text{G}=
       CD\left(\mathcal A_{H_G}, \mathcal A_{O_G}\right)
\end{equation}

\subsubsection{\textbf{Functional Part Distance ($\boldsymbol{d_\text{F}}$)} for Functionality}
Measures the alignment between the hand's functional part $\mathcal A_{H_F}$ and the object's functional part $\mathcal A_{O_F}$:

\begin{equation} \label{eq:df}
    d_\text{F}=\min CD\left(\mathcal A_{H_F}, \mathcal A_{O_F}\right) 
\end{equation}

\subsubsection{\textbf{Hand-Object Interpenetration Depth ($\boldsymbol{d_\text{IP}}$)} for Penetration}
Defined as the most negative signed distance field (SDF) from $\mathcal{P_H}$ to $\mathcal{O}$, indicating maximum intrusion.

\begin{equation}
    d_\text{IP} = - \min\left[\min SDF\left(\mathcal{P_H} | \mathcal{O}\right), 0\right]
\end{equation}

\subsubsection{\textbf{Hand Self Penetration Depth ($\boldsymbol{d_\text{SP}}$)} for Penetration}
Measures the maximum penetration between different hand parts $\mathcal H_{p_i}$ and $\mathcal H_{p_j}$, where $\mathcal P_{H_{p_i}}$ denotes surface point cloud:
\begin{equation}
    d_\text{SP} = - \min\left[\min \mathop{SDF}_{i \neq j}\left(\mathcal P_{H_{p_i}} | \mathcal H_{p_j}\right), 0\right]
\end{equation}

\begin{table}[!tbp]
\caption{Evaluation Across Different Data Volumes}
\label{tab:net_volume}
\renewcommand{\arraystretch}{1.2}
\centering
\resizebox{\linewidth}{!}{
\begin{threeparttable}
    \begin{tabular}{c|c|c|c|cc}
        \Xhline{1.2pt}
        \multirow{2}{*}{Training set$^*$} &
         
          Stability & Contact &
          Functionality &
          \multicolumn{2}{c}{Penetration} \\ \cline{2-6} 
         
           &
          $SR\,\uparrow$ &
          $d_\text{G} / \text{cm}\,\downarrow$ &
          $d_\text{F} / \text{cm}\,\downarrow$ &
          $d_\text{IP} / \text{cm}\,\downarrow$ &
          $d_\text{SP} / \text{cm}\,\downarrow$ \\ \hline
          \begin{tabular}[c]{@{}c@{}}
          0.1k grasps / Cat.\\ \textcolor{gray}{($>$ DexFG\cite{dexfg} volume)}\end{tabular}
         &
          
           53.2\%&
          1.926 &
          2.259 &
          2.272 &
          0.164 \\
        \begin{tabular}[c]{@{}c@{}}0.3k grasps / Cat.
        \end{tabular}& 
         59.8\%&
          1.818 &
          1.885 &
          1.931 &
          0.202 \\

        \begin{tabular}[c]{@{}c@{}}0.5k grasps / Cat.
        \end{tabular}&  
         61.3\%&
          1.808 &
          1.626 &
          1.800 &
          0.202 \\
          
        \begin{tabular}[c]{@{}c@{}}1k grasps / Cat. 
        \end{tabular}&  
          62.7\%&
          1.796 &
          1.544 &
          1.745 &
          0.212 \\
        \begin{tabular}[c]{@{}c@{}}3k grasps / Cat. \end{tabular} 
          & 65.2\% & 1.717  & 1.253  & 1.443  & 0.230  \\
        \begin{tabular}[c]{@{}c@{}}10k grasps / Cat. \end{tabular}
          & \textbf{75.0\%} & \textbf{1.673}  & \textbf{1.054}  & \textbf{1.123}  & \textbf{0.137}  \\
        \Xhline{1.2pt}
    \end{tabular}
    \begin{tablenotes}
    \footnotesize
    \item All data were tesed on an identical \textbf{0.1k grasps / Cat.} test set. 
    \item All data is trained until convergence and has the same number of training steps.
    \end{tablenotes}
    \vspace{-5mm}
    \end{threeparttable}
}

\end{table}

\begin{table}[!bp]
\vspace{-2mm}
\vspace{-4mm}
    \caption{Comparison of ScaleADFG Net on Generated (ScaleADFG) \\and Real Objects Dataset (DexFG)}
    \label{tab:net_ablation}
    \renewcommand{\arraystretch}{1.2}
    \resizebox{\linewidth}{!}{
        \begin{tabular}{c|c|c|c|cc}
        \Xhline{1.2pt}
        \multirow{2}{*}{Dataset} &
        Stability & Contact &
        Functionality &
        \multicolumn{2}{c}{Penetration} \\
        \cline{2-6}
        & $SR\,\uparrow$ 
        & $d_\text{g} / \text{cm}\,\downarrow$ & $d_\text{f} / \text{cm}\,\downarrow$ & $d_\text{IP} / \text{cm}\,\downarrow$  & $d_\text{SP} / \text{cm}\,\downarrow$ \\
        \hline
            ScaleADFG dataset & 76.7\% & 1.673 & 1.054 & 1.123 & 0.137  \\
            DexFG dataset\cite{dexfg} & 75.0\% & 1.934 & 1.213 & 1.868 & 0.238 \\ \Xhline{1.2pt}
        \end{tabular}
    }
    \vspace{-2mm}
\end{table}

\subsection{ScaleADFG Dataset Statistics}

\subsubsection{Dataset Comparison}

We compare ScaleADFG dataset with similar datasets, as summarized in \cref{tab:dataset_comparison}, suggesting ScaleADFG is among the first datasets of dexterous functional grasps at a large scale across multiple scale variations.

\subsubsection{Dataset Statistics}  

The ScaleADFG dataset comprises 5 object categories and 2 robotic hands (Allegro and Shadow). Each category contains over 1,000 objects, with more than 10,000 functional grasps per category per hand. 

\begin{figure}[!tbp]
    \centering
    \includegraphics[width=0.85\linewidth]{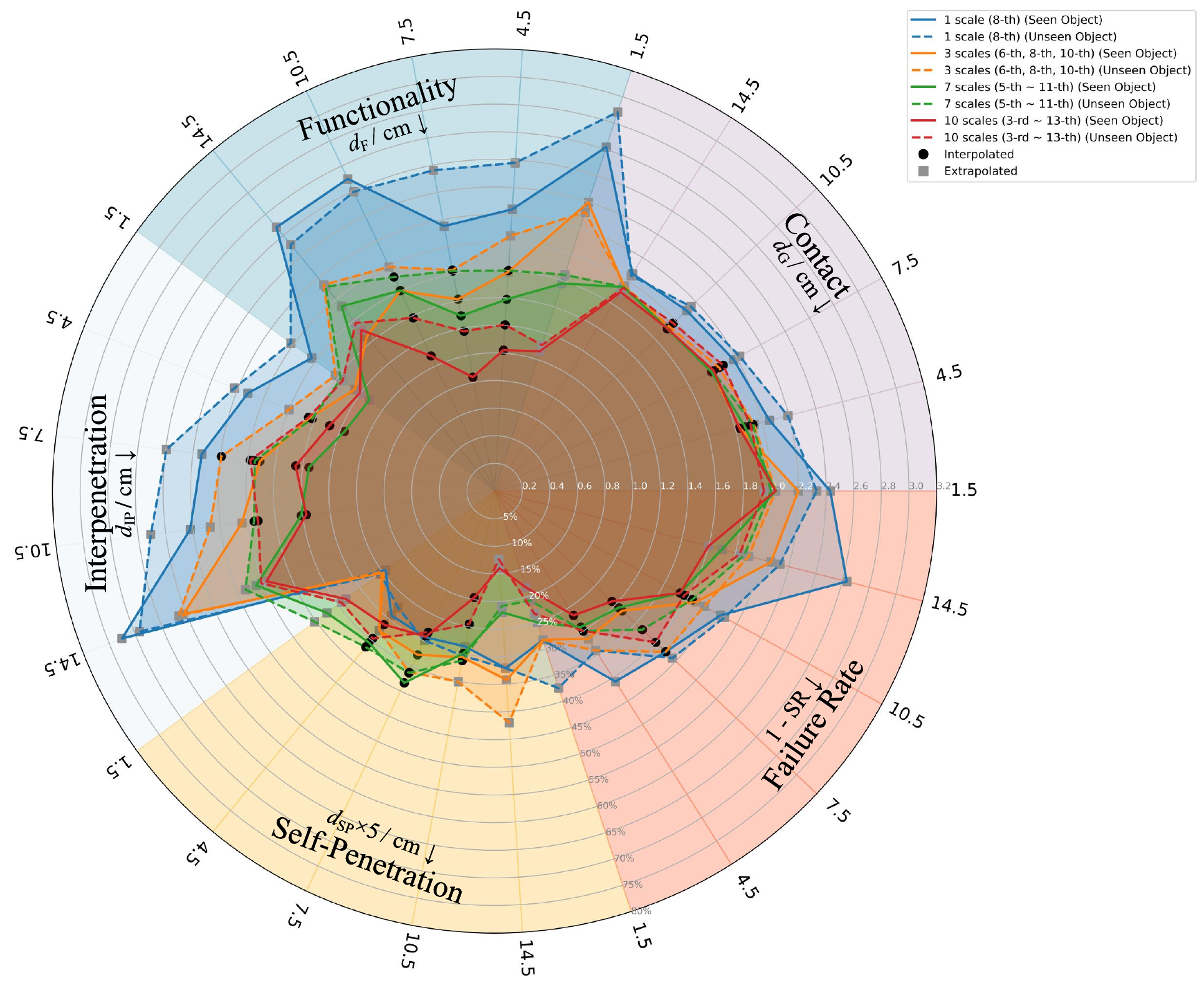}
    \caption{\textbf{Evaluation of Scale Generalization Performance.} All data were tested on \textbf{Unseen Scale} both on seen objects and unseen objects. Unseen scales were interpolated from existing scales in the dataset. The interpolation between the 1st and 2nd scales is referred to as the 1.5th scale. Testing was conducted on five unseen scales: 1.5, 4.5, 7.5, 10.5, and 14.5, with both extrapolated scales and interpolated scales. Besides, since all metrics are better when smaller, a \textbf{smaller area} is preferable.
}
    \label{fig:net_scale}
    \vspace{-2mm}
\end{figure}
\subsubsection{Dataset Quality}
A human study (50 objects per category, 3 participants) was conducted to assess affordance retrieval accuracy, as shown in \cref{tab:human_study}. Affordance accuracy is computed as $\frac{|\hat{P}\cap P_{\text{gt}}|}{|\hat{P}|}$, where $\hat{P}$ and $P_{\text{gt}}$ denote retrieved and human-labeled affordance points, respectively.
\begin{figure}[!bp]
    \vspace{-4mm}
    \centering
    \includegraphics[width=1.0\linewidth]{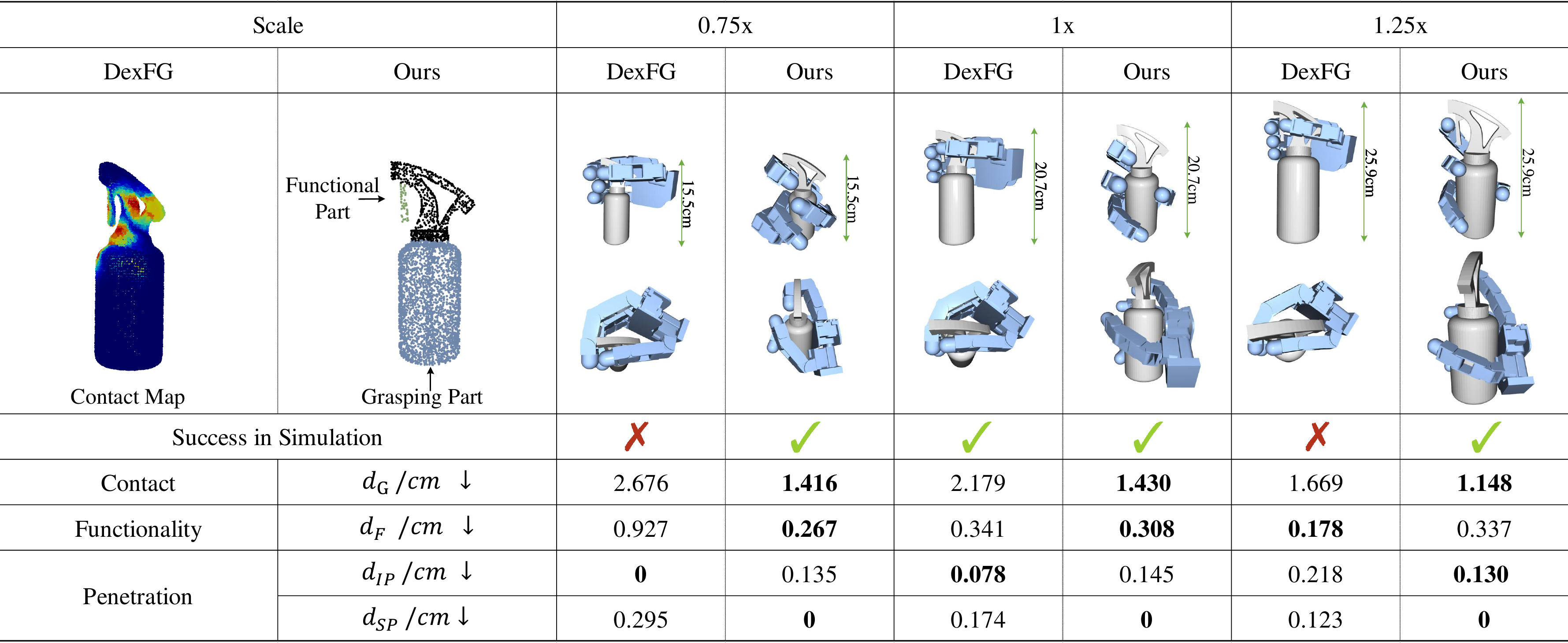}
    \caption{ Comparison with DexFG \cite{dexfg} on functional grasping across different object scales, demonstrating the enhanced scale-adaptability of ScaleADFG. \hl{For a fair comparison, we define the entire object body as the grasping part for both methods during evaluation.}
    }
    \label{fig:compare_dexfg_scale_affordance}

\end{figure}

\subsection{Effect of Object Scale Diversity and Dataset Volume on Grasp Prediction}
\subsubsection{Effect of Object Scale Diversity}
\cref{fig:net_scale} quantitatively evaluates grasp performance across different training scale settings on both seen and unseen objects with previously unseen scales. The results demonstrate that the model generalizes effectively to unseen scales, with competitive performance on both interpolated and extrapolated scales. Increasing the diversity of training scales consistently enhances grasp quality for both seen and unseen objects. Across all configurations, grasp quality on unseen objects remains comparable to that on seen objects, indicating robust generalization across scales.

\subsubsection{Effect of Dataset Volume on Grasp Prediction}
Results in \cref{tab:net_volume} indicate that increasing the volume of training data significantly enhances the performance of grasp prediction across stability, contact, functionality, and penetration metrics. This highlights the importance of large-scale datasets, such as ScaleADFG, for training more robust and efficient models.

\subsubsection{Evaluation of ScaleADFG Net on Transferring from Generated to Real Object} Results in \cref{tab:net_ablation} show the model performs slightly less well on the real objects (DexFG \cite{dexfg} dataset) compared to the generated objects (our ScaleADFG dataset), but the results remain largely comparable. 

\subsection{Comparison with Other Methods}

\begin{figure}[!tbp]
    \centering
    \includegraphics[width=\linewidth]{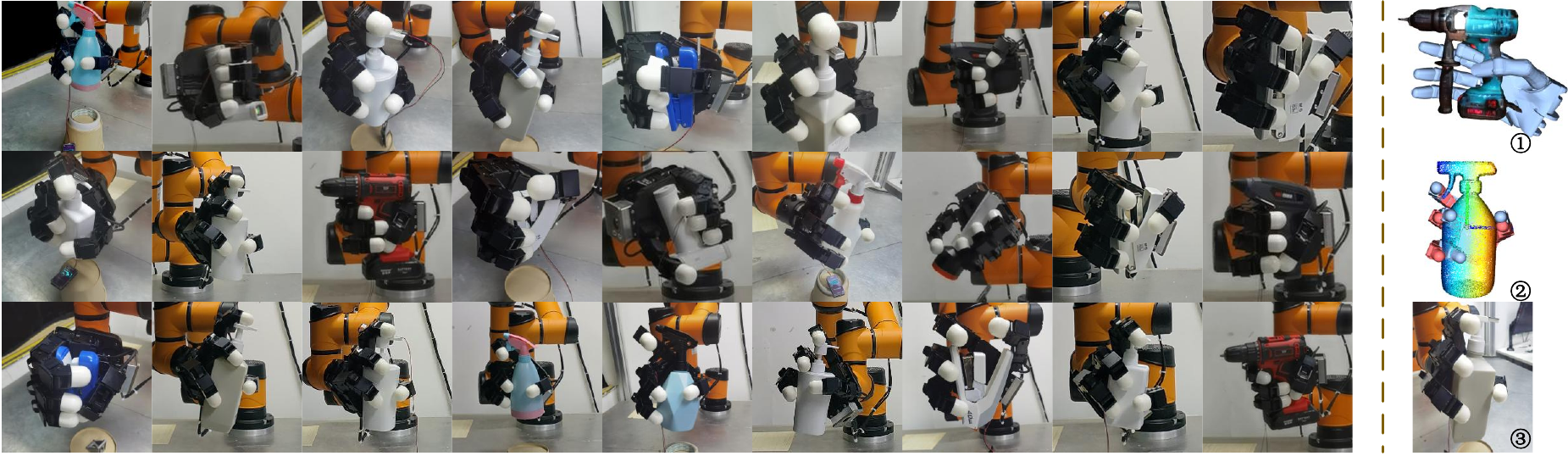}
    \caption{Functional grasping with the Allegro Hand on a real-world platform is shown on the left. \hl{On the right are representative failure cases: (1) during optimization, the narrow spacing between the two handles prevents synthesis of collision-free grasps, leading the hand to grasp outside both handles; (2) during network inference, the predicted grasp may induce slight collisions (prediction in blue, ground truth in red); (3) on the real-world platform, object motion during grasping causes the functional finger contact to slip off the object’s functional part.} }
    \label{fig:experiments}
    \vspace{-4mm}
\end{figure}

\begin{figure}[!bp]
    \vspace{-5mm}
    \vspace{-4mm}
    \hspace{-3.5mm}
    \subfloat[Real-world platform]{
        \label{fig:platform}
        \includegraphics[width=0.42\linewidth]{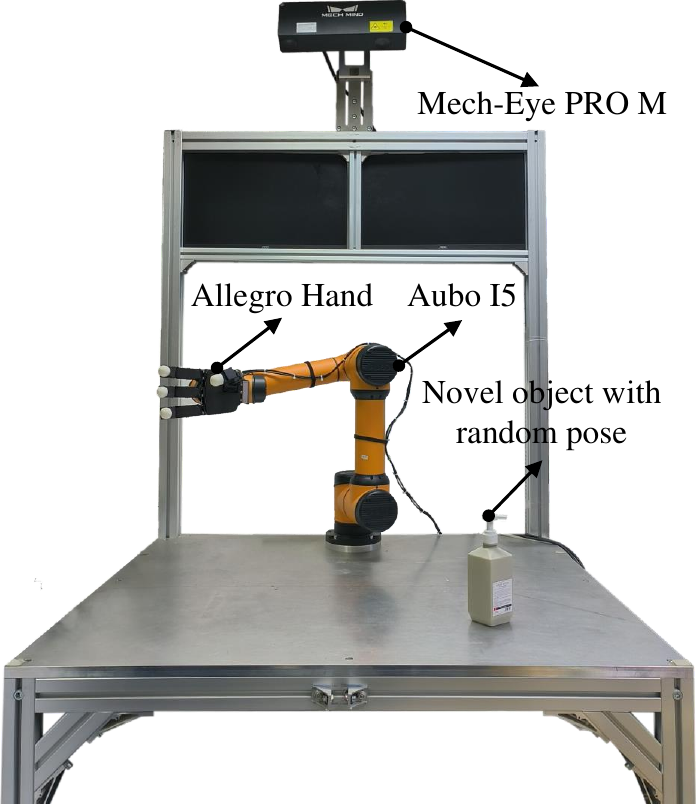}
    }
    \hspace{-3.5mm}
    \subfloat[List of objects]{
        \label{fig:object}
        \includegraphics[width=0.41\linewidth]{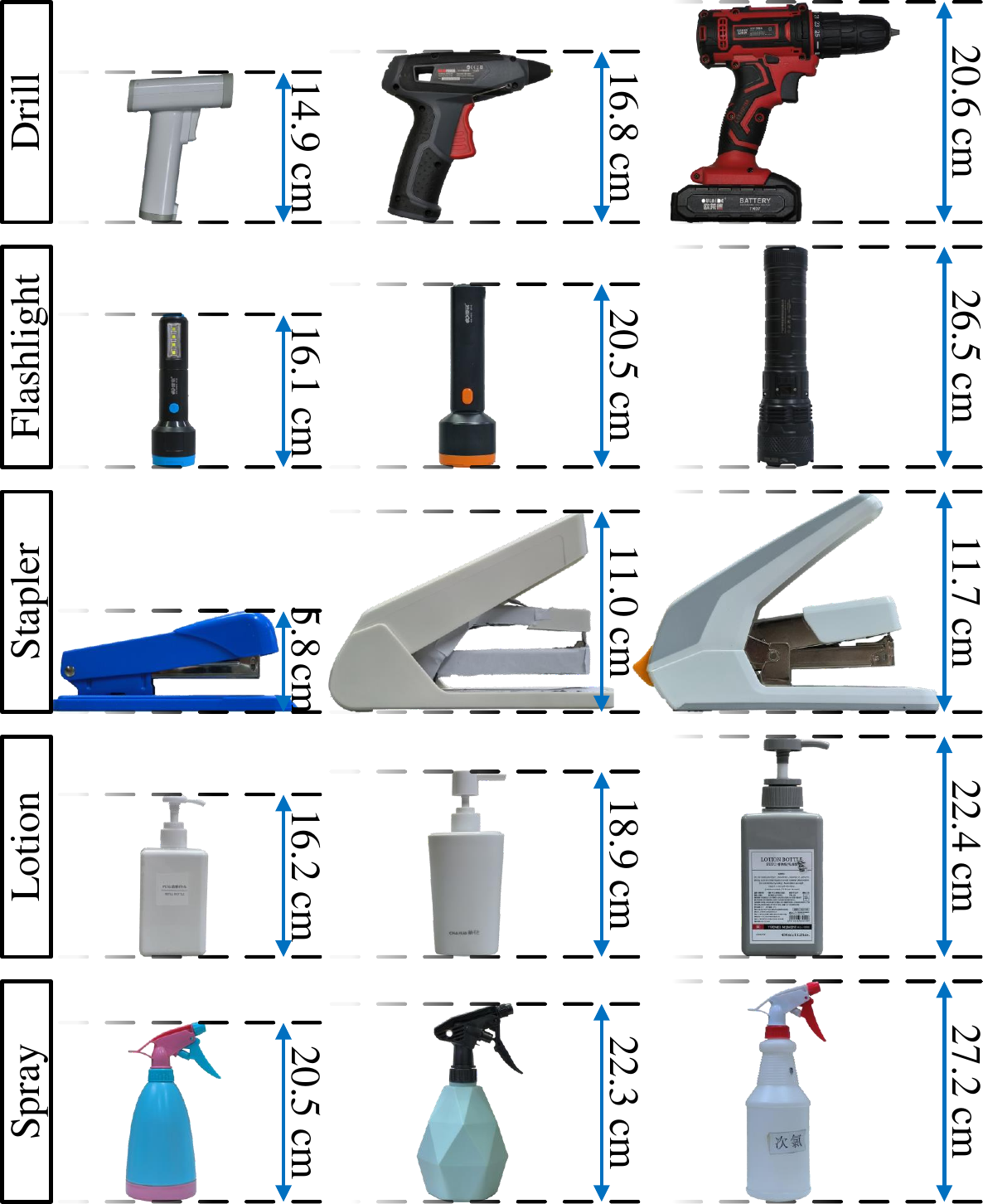}
    }
    \makebox[0pt][l]{\hspace{-4.5mm} 
    \subfloat[\hl{Sensors}]{
        \label{fig:force_sensors}
        \includegraphics[width=0.16\linewidth]{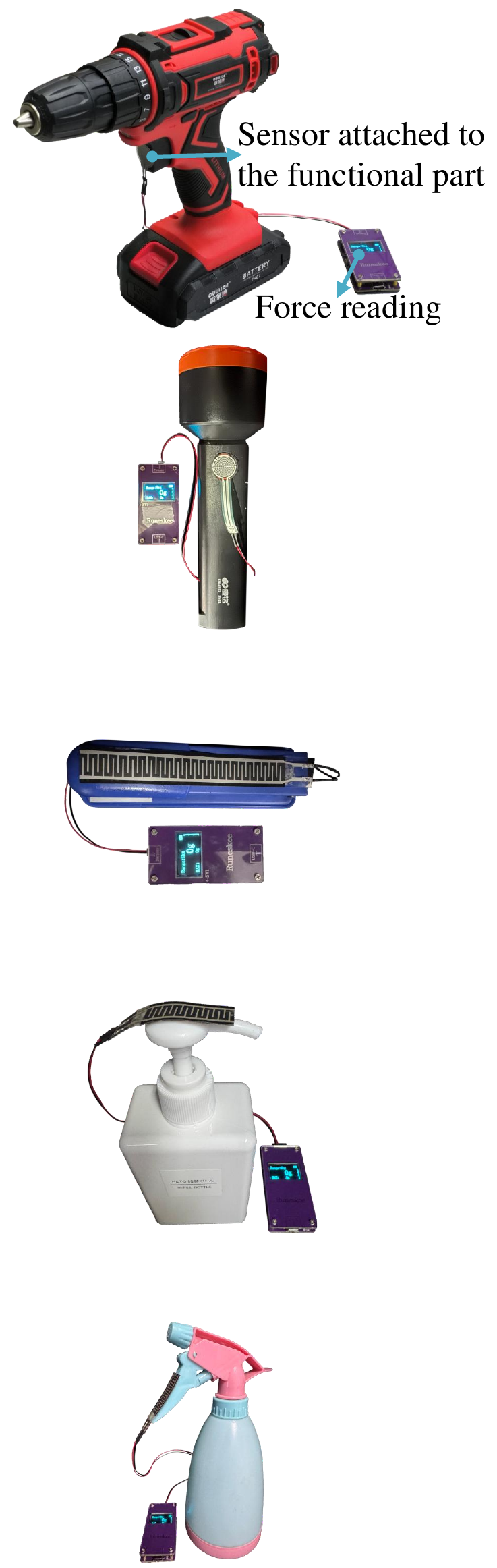}
    }}
    \label{fig:real}
    \caption{Real-world experimental setup. (a) Aubo I5 robotic arm equipped with an Allegro Hand and a Mech-Eye PRO M camera for physical grasp execution. (b) A diverse set of 15 novel objects used in the experiments, with size markings shown on the side. \hl{(c) Force sensors attached to the functional part. To accommodate varying affordance area geometries across object categories, force sensors of different sizes and shapes are employed.}}
\end{figure}

\subsubsection{Grasp Synthesis Across Varying Scales}
We compare ScaleADFG with DexFG \cite{dexfg} on objects at different scales $(0.75 \times, 1\times, 1.25\times)$. As shown in \cref{fig:compare_dexfg_scale_affordance}, ScaleADFG achieves more consistent contact and lower interpenetration across scales, while DexFG suffers from limited adaptability due to rigid contact map constraints. Our method also demonstrates reduced self-penetration and superior functional contact, especially for small-scale objects. Besides, in DexFG \cite{dexfg}, all grasps are restricted to the upper part of the spray bottle based on its contact map, which differs from our definition of the grasping part. For a fair comparison, we designate the entire object body as the grasping part for both methods during evaluation.

\subsubsection{Grasp Prediction Quality}
To further strengthen the evaluation, we compared our model (trained on Shadow Hand) as \cref{tab:sr} with DexFuncGrasp \cite{hangDexFuncGraspRoboticDexterous2024} and FunctionalGrasp \cite{zhang2023functionalgrasp} on common categories, according to DexFuncGrasp's \cite{hangDexFuncGraspRoboticDexterous2024} official simulation code and object sizes, as shown in \cref{tab:sr}.

\subsection{Experiments on Real Robot}
Real-world experiments are conducted using an Aubo I5 robotic arm, Allegro Hand, and Mech-Eye PRO M camera (\cref{fig:platform}). A diverse set of 15 novel objects, varying in category and scale, are randomly placed for testing. We validate both grasp stability and functionality in real-world experiments.

\subsubsection{Experiment pipeline}

\begin{table}[!tbp]
    \caption{Comparision of Success Rate}
    \label{tab:sr}
    \renewcommand{\arraystretch}{1.2}
    \centering
    \resizebox{1.0\linewidth}{!}{
    \begin{threeparttable}
        \begin{tabular}{l|cccc|c}
            \Xhline{1.2pt}
            & Drill              & Flashlight           & Stapler           & Spray Bottle  & Avg. \\
            \hline
            FunctionalGrasp \cite{zhang2023functionalgrasp}     & -                 & 31.25\%               & \textbf{100\%}     &  -            & -    \\
            DexFuncGrasp \cite{hangDexFuncGraspRoboticDexterous2024}            & 55.00\%           & 91.03\%                & 85.77\%          &  58.07\%      & 72.47\%     \\
            ScaleADFG (Ours)                                                & \textbf{100\%}     & \textbf{91.67\%}       &  96.67\%        & \textbf{90.00\% }      & \textbf{94.58\%}   \\
            \Xhline{1.2pt}
        \end{tabular}
    \begin{tablenotes}
        \footnotesize
        \item Baseline methods did not train on dexterous functional grasp dataset with \textbf{scale variety}. For a fair comparison, we compare our method on objects in their test dataset without change object scale.
    \end{tablenotes}
    \end{threeparttable}
    }
     \vspace{-4mm}
\end{table}

We conduct real-world experiments using the Iterative Closest Point (ICP) registration algorithm \cite{rusinkiewicz2001efficient_icp} in conjunction with offline-generated object models. Initially, we capture 4 images from different perspectives around the object to be grasped. These images are then segmented using Grounded-SAM 2 \cite{renGroundedSAMAssembling2024, ravi2024sam2segmentimages} and sent to TRELLIS \cite{xiang2024trellis} for 3D reconstruction. The resulting surface points are employed for ICP registration to determine the object's 6D pose in the real-world experiment. Finally, the grasp configuration inferred by the ScaleADFG network is transformed to the world coordinate and executed. Specifically, we apply a 0.2 radians adjustment to the flexion of inward-bending joints to enhance grasp stability.

\subsubsection{Experiment results}

As shown in \cref{fig:experiments}, each object is tested in 5 trials at different poses. The Grasp Success Rate ($SR$) is defined as the proportion of successful grasps (object held for more than 10 seconds), \hl{$SR_\text{func}$ is computed over the same number of trials as $SR$, inspired by} \cite{zhu2023toward}, \hl{but only considers grasps successful under the $SR$ criterion that further achieve a nonzero force reading, indicating functional task execution (e.g., button activation).}
The system achieves $SR=76.0\%$ and $SR_\text{func}=58.7\%$ across all objects. Performance is consistent across object scales, indicating strong generalization and scale invariance.

\begin{table}[!bp]
\vspace{-2mm}
\vspace{-4mm}
    \caption{Success Rates in Real-World Experiments}
    \label{tab:real_robot}
    \renewcommand{\arraystretch}{1.2}
    \resizebox{\linewidth}{!}{
    \begin{threeparttable}
        \begin{tabular}{c|c|ccccc|c}
            \Xhline{1.2pt}
            Scale & Metrics & Drill & Flashlight & Stapler & Lotion bottle & Spray bottle & Avg. \\
            \hline
            \multirow{2}{*}{S} & $SR$ & $4/5$ & $3/5$ & $4/5$ & $4/5$ & $5/5$ & 80.0\% \\
                               & $SR_\text{func}$ & $2/5$ & $2/5$ & $3/5$ & $3/5$ & $4/5$ & 56.0\% \\
            \cline{2-8}
            \multirow{2}{*}{M} & $SR$ & $3/5$ & $2/5$ & $3/5$ & $5/5$ & $5/5$ & 72.0\% \\
                               & $SR_\text{func}$ & $3/5$ & $1/5$ & $2/5$ & $4/5$ & $4/5$ & 56.0\%  \\
            \cline{2-8}
            \multirow{2}{*}{L} & $SR$ & $4/5$ & $3/5$ & $4/5$ & $4/5$ & $4/5$ & 76.0\% \\
                               & $SR_\text{func}$ & $4/5$ & $2/5$ & $4/5$ & $3/5$ & $3/5$ & 64.0\% \\
            \hline
            \multirow{2}{*}{Avg.} & $SR$ & 73.3\% & 53.3\% & 73.3\% & 86.7\% & 93.3\% & 76.0\% \\
                               & $SR_\text{func}$ & 60.0\% & 33.3\% & 60.0\% & 66.7\% & 73.3\% & 58.7\% \\
            \Xhline{1.2pt}
        \end{tabular}
    \begin{tablenotes}
        \footnotesize
        \item ``S'' denotes small objects, ``M'' denotes medium-sized objects, and ``L'' denotes large objects.
    \end{tablenotes}
    \end{threeparttable}
    }
\end{table}

\section{Conclusion}

We propose the ScaleADFG pipeline, highlighting both the scalability of dexterous functional grasp dataset construction and adaptability across diverse object scales. Our contributions span two areas: algorithm design and dataset creation. We introduce a large-scale dataset specifically designed to capture scale variability, addressing the common bias in existing datasets toward oversized objects due to robotic hand size constraints. On the algorithmic side, we develop an affordance-based grasp generation method that overcomes the limitations of existing contact map approaches, which are sensitive to hand-object scale ratios. Our pipeline is enhanced for automation and efficiency, leveraging pretrained models to minimize manual annotation and enhance both the efficiency and generalizability of dexterous functional grasp generation.

\hl{\textbf{Future Work.} Our framework already generalizes well to rigid objects and button-press articulated tools, showing strong adaptability to scale and geometry variations. Building on this capability, we aim to extend it toward deformable objects and more complex articulated tools (e.g., scissors), where dynamic affordance predictions and fine-grained contact modeling will be required. The modular design of our pipeline makes such extensions feasible, including incorporating deformable object simulation and learning-based affordance discovery. Moreover, although our network demonstrates robustness to moderate noise and occlusion, extending it to cluttered real-world scenes with heavy occlusion is a natural next step. Finally, moving beyond functional grasp generation to subsequent fine-grained manipulation after grasping will further broaden the scope and impact of dexterous functional grasping.}

\bibliographystyle{IEEEtran}

\bibliography{references}

\end{document}